ORIGINAL ARTICLE

# Effects of data time lag in a decision-making system using machine learning for pork price prediction

**Mario E. Suaza-Medina[1] · F. Javier Zarazaga-Soria[1] · Jorge Pinilla-Lopez[1] · Francisco J. Lopez-Pellicer[1] · Javier Lacasta[1]**

**Abstract**

Spain is the third-largest producer of pork meat in the world, and many farms in several regions depend on the evolution of this market. However, the current pricing system is unfair, as some actors have better market information than others. In this context, historical pricing is an easy-to-find and affordable data source that can help all agents to be better informed. However, the time lag in data acquisition can affect their pricing decisions. In this paper, we study the effect that data acquisition delay has on a price prediction system using multiple prediction algorithms. We describe the integration of the best proposal into a decision support system prototype and test it in a real-case scenario. Specifically, we use public data from the most important regional pork meat markets in Spain published by the Ministry of Agriculture with a two-week delay and subscription-based data of the same markets obtained on the same day. The results show that the error difference between the best public and data subscription models is 0.6 Euro cents in favour of the data without delay. The market dimension makes these differences significant in the supply chain, giving pricing agents a better tool to negotiate market prices.

## 1 Introduction

The agricultural sectors are essential in most economies of the world. Therefore, controlling the prices of their products is crucial, as they have a significant impact on financial markets, affecting supply and demand and directly influencing the entire supply chain, from producers to final consumers.

Price volatility is inevitable [35], and it can create uncertainty in the markets [4], impacting the economy of producers and the country as a whole. Sometimes, it indicates poor market behaviour, while at other times, it is caused by the natural seasonality of agricultural products.

For producers, projecting the future behaviour of associated markets is vital in reducing production risks and facilitating product management. Knowing future price estimates of their products allows them to design appropriate strategies to address price volatility. In the case of governments, early price forecasting helps them understand the behaviour of the country's economy, facilitating the creation of policies for the management of economic resources, as well as preparation for different scenarios that may arise.

The European Union (EU) recognizes the importance of agriculture market management by making an exhaustive registry of the prices of the main products in European countries. This is created using the information provided by member states that periodically inform the European Commission about the reference prices of the agricultural and livestock products in their countries. The Statistical Office Of The European Communities (Eurostat) uses this information to provide monthly public reports with weekly prices of these products. This information is used to supervise the shared organization of European agriculture, including the direct payment to farmers under support schemes [11, 12].

These European markets operate through agreements between sellers and buyers based on reference prices usually fixed for regions or countries. The processes for determining these prices vary, but many markets still have manual price decision processes based on finding consensus among the main agents. These agents make proposals based on market interpretation, private information about their business processes, and public market information. These proposals, acting as a market prediction, help set the final price. However, this manual price agreement process is only suitable for some of the involved actors because those with less information about the state of the market are at a disadvantage in price negotiations.

✉ Mario E. Suaza-Medina
mesuaza@unizar.es

[1] Universidad de Zaragoza, Zaragoza, Spain

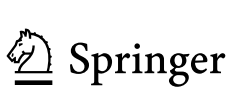

In this context, a decision support system that provides information about the expected price evolution according to market factors can help make better-informed and fairer decisions. However, automatic price prediction presents multiple challenges:

- Small margin of error. A slight price variation can result in a significant business loss due to the large amount of product traded.
- Demand is volatile, although some behavioural patterns can be found [21].
- Many unpredictable/unquantifiable factors, such as diseases, weather, or economic changes, can suddenly affect the price of a product.

Historical pricing is easy to find, and affordable data can be used to construct such decision support. However, the time lag in the data acquisition can affect its price decision. The further in time the most recent data in the series, the higher the probability of unpredictable factors affecting the price.

In this paper, we study the effect that data time lag has on the price predictions of pork prices in the Spanish regional market of Lleida. Specifically, Spain is the third producer of pork meat after China, and the USA,[1] and the area surrounding Lleida accounts for more than half of the Spanish pork production. This is a big market whose price is affected by many internal and external factors.

For predicting the prices, we have analysed the historical behaviour of other European markets with available information concerning Lleida to select the most suitable ones. Concerning the historical data used, we have considered two types: data that will be obtained from direct sources of public information, which have a publication time delay and will be referenced as public data, and private data that can be obtained without a time delay through a paid subscription, which will be referenced as subscription data. In subscription data, we use that the weekly market prices are not set simultaneously but scaled along weekdays. Since the markets that publish their weekly prices earlier have indirectly dealt with uncertainties when selecting their reference prices, they indicate price trends for the chosen market.

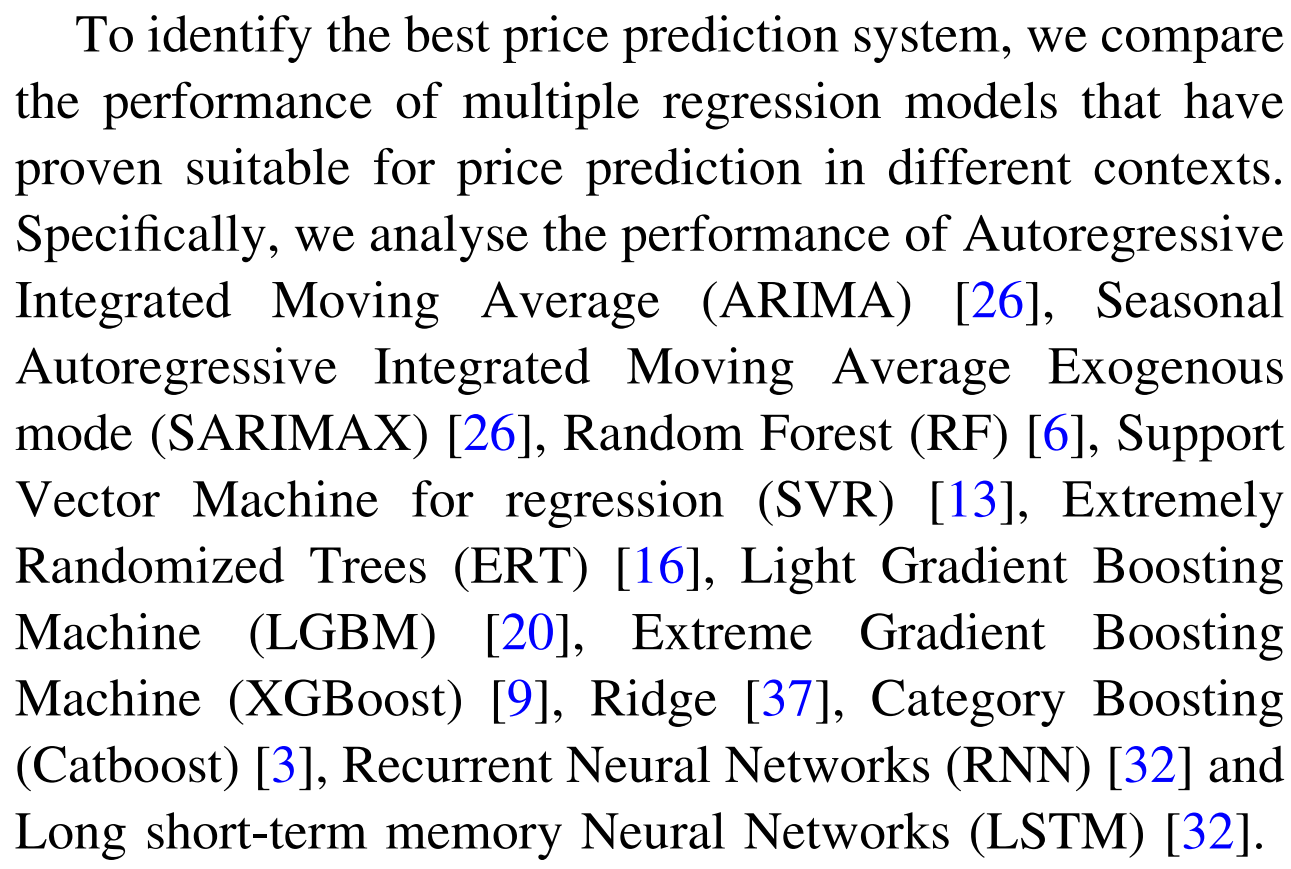

To identify the best price prediction system, we compare the performance of multiple regression models that have proven suitable for price prediction in different contexts. Specifically, we analyse the performance of Autoregressive Integrated Moving Average (ARIMA) [26], Seasonal Autoregressive Integrated Moving Average Exogenous mode (SARIMAX) [26], Random Forest (RF) [6], Support Vector Machine for regression (SVR) [13], Extremely Randomized Trees (ERT) [16], Light Gradient Boosting Machine (LGBM) [20], Extreme Gradient Boosting Machine (XGBoost) [9], Ridge [37], Category Boosting (Catboost) [3], Recurrent Neural Networks (RNN) [32] and Long short-term memory Neural Networks (LSTM) [32].

The main contributions of the paper are the following:

- Analysing the available market information about pork prices in European countries and identifying the existing correlations.
- Study how the data time lags of the data affect the quality of the predictions.
- Compare statistical, machine learning and deep learning solutions for pork meat price prediction. This allows for determining if some technique is especially suitable for the task or if several can be indistinctly selected.
- Describes a prototype that integrates the best technique in a weekly market price proposal.

The paper is organized as follows. The next section describes state of the art in price prediction techniques. Then, we analyse the available data sources with pork price information in Europe and select those suitable for the experiments. Next, we describe the experiment details and parameters of the previously described prediction techniques and compare the results. After that, we describe the prototype developed to use the best-identified technique in a weekly market price proposal. The paper concludes with conclusions and describes the current limitations and future work.

## 2 State of the art

The literature provides multiple techniques to forecast livestock, pork and crop markets. The proposed solutions have evolved from mathematical and statistical models to machine learning approaches that combine statistical and artificial intelligence models to provide better predictions. Among these models, Arima, Sarimax, Bayesian networks, Random Forests, Support Vector Machines, and Neural networks are the most popular, but other models have also been used.

Miah et al. [24] designed a generic environment to support the construction of decision support systems (DSS). At that time, adopting application systems in the rural environment was scarce. Therefore, this research proposes a DSS that considers contextual factors, which are important for the end user to achieve this and involves experts in creating the basis of the system as end users who are the business operators. As a result, the end-users have a customisable system where each specific decision system presents expert suggestions to help producers in their farming practices. Shih et al. [33] work proposes a weighted case-based reasoning (CBR) approach to construct a price prediction model for broilers using features related to the price and quantity of chicks, pigs, feedstuff, and price indexes. It utilizes past occurrences of similar market conditions to infer the current price of broilers. Zhou and Pei [44] introduce the Generalized Grey Verhulst Model (GGVM) to know the price index of pigs, and its results indicate that the GGVM gives better simulations and more accurate prediction than least square support vector regression, support vector regression and radial basis function networks. Meanwhile, [31] describes a system based on artificial intelligence to manage resources efficiently. This system allows setting prices for different products and services independent of the demand model used and generating guidelines for pricing and production. This lowers the cost of implementation of revenue management systems because the user only builds a model but not a specific algorithm for each problem.

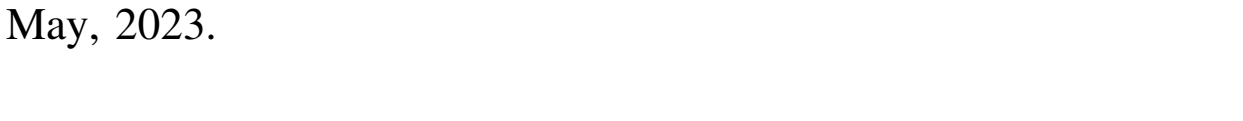

[1] https://www.pig333.com/pig-production-data/graficos/, Last visit May, 2023.

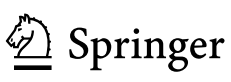

To forecast pig prices, [40] proposes a smoothing generator operator that uses the history of prices to predict future prices. This operator extends the double exponential smoothing method with a grey accumulating generation operator that can smooth the random interference of data. Ahumada and Cornejo [1] identify the strong relations between different markets and propose to improve the accuracy of individual food price models by considering the cross-dependence between them. They use an equilibrium correction model to find the residual cross-correlations of corn, soybeans, and wheat prices. Tao et al. [34] use an empirical ensemble mode decomposition to extract high, low, and trend frequency behavioural patterns in the evolution of hog prices. Frequency patterns are then analysed using an ARIMA, an extreme learning machine, a neural network, and a polynomial function. Xiong et al. [41] propose using an ensemble for vegetable price prediction that separates seasonal and trend components using decomposition procedures based on losses. They use these trends as input for extreme learning machines for short, medium, and long prediction.

Wang [38] proposes a web platform that integrates essential services in agricultural intelligence computing, farm production tracking, purchase–sales–storage, enterprise management, live pig trading, and supply chain finance. The work includes a price forecasting service for pig prices using an ARIMA model. Paroissien [25] compares the performances of univariate Unobserved Component Models (UCM), Holt-Winters Exponentially Weighted Moving Average (EWMA), Vector Error Correction Model (VECM) and Vector Auto-Regressive (VAR) forecast models to predict the price of Bordeaux wine based on the yearly and monthly price evolution of 15 different wine types. Pavlyshenko [26] studies using machine-learning models for sales prediction in Rossmann stores. They propose to build a set of individual regression models, using stacking techniques with which he seeks to improve the performance of the models using small datasets. Liu etal. [21] identified the pseudo-cyclical fluctuations of the hog prices in two Chine provinces. They use a Hodrick–Prescott filter to separate the pseudo-cyclical fluctuations from the price trends caused by external factors.

Machine Learning applications in the pork sector began to strengthen in 2018. The principal models used were neural networks and random forest algorithms. Ma et al. [23] compare the performance of different price prediction models for pork using features related to other livestock prices, feedstuff, and market indexes. They compare the results of a dynamic Bayesian network, a support vector machine, and a fully connected neural network with the classic ARIMA method. Shahinfar et al. [29] propose models to infer animal features that affect pricing. This work assumes the hot carcass weight, the intramuscular fat, the Greville rule fat depth, the computed tomography lean meat yield, and the loin weight based on phenotypic and environmental predictor features. They compare the performance of a deep neural network, a Gradient boosting tree, a K-nearest neighbour solution, a model tree, and a random forest. They continued this work in [30] by inferring beef traits relevant to market pricing. They use features such as age, weight, and body dimensions and compare using a model tree, a random forest, a support vector machine model, and a neural network. They used a synthetic oversampling minority approach. Again [22] designed a network service platform with which they could obtain the feeding status of each pig in real-time, the identification, health, and the exact quantity to be allocated. It was also found that the container in which the feed was kept was the factor that most influenced feeding. Previous research did not consider the pigs' living environment and physical condition. With the accumulation of data generated daily, it is proposed to use machine learning algorithms to improve decision-making and to know the feeding status of the pigs.

Ribeiro [28] propose predicting soybean and wheat prices using historical prices. The paper compares a Gradient boosting machine, an Extreme Gradient boosting machine, a Random Forest, Support vector regressors, and a K-nearest neighbours algorithm. Chuluunsaikhan et al.

[10] propose a methodology that predicts the daily retail price of pork in the South Korean domestic market based on news articles. They use topic modelling techniques that obtain relevant keywords that can explain the price fluctuations with these words. They apply machine learning, statistical and deep learning methods as part of the process. Zhang et al. [43] propose a solution for price pork forecasting, using a hybrid model that, instead of making specific forecasts, suggests price intervals since it is believed to be more effective and helpful as it provides more information about the data generation process. They compare this model with other single and hybrid models, obtaining better results with their proposal. Ye et al. [42] build a Heterogeneous Graph-enhanced LSTM (HGLTSM) to predict weekly hog prices using historical prices and discussions from the online professional community to create heterogeneous graphs. The findings show that forum information is beneficial to hog price prediction.

In recent works, [39] describe a WOA-LightGBM-CEEMDAN model for hog price forecasting. The Whale Optimization Algorithm (WOA) is used to optimise the LightGBM (Light Gradient Boosting Model) parameters and the residual sequence is decomposed and corrected using the CEEMDAN model (Complete Ensemble Empirical Mode Decomposition with Adaptive Noise). Klompenburg [36] build a literature review on data-driven decision-making on pig farms, mainly focusing on reviewing real business cases where the data source comes from real production chains. The results found that there are few real cases, and the vast majority are experimental with small instances of animals, in other words, in controlled scenarios and not real life. Chen et al. [8] develop a combination of neural network models based on bidirectional RNN and bidirectional LSTM for price prediction. The results found that with the increase in training, the predictions improve, but in the case of too many, the values do not improve and will start to decrease. The simple structure of unidirectional and bidirectional models has a large prediction margin, but better price predictions can be obtained by combining them. Punia and shankar [27] propose designing a decision-making system based on deep learning and proposing an algorithm for real-time demand forecasting, combining LSTM networks and random forests to obtain good short- and long-term results. The data used are for food products. Finally, [19] proposes network self-attention to extract more information from time series, proposes a new RNN to learn the similarity between nodes in a random process, the similarity scores are normalized and calculated by weighted summing by softmax in their results, the proposed forecasting method indicates that network self-attention has a high ability to measure the similarities of nodes in the network and make best scores in the predictions.

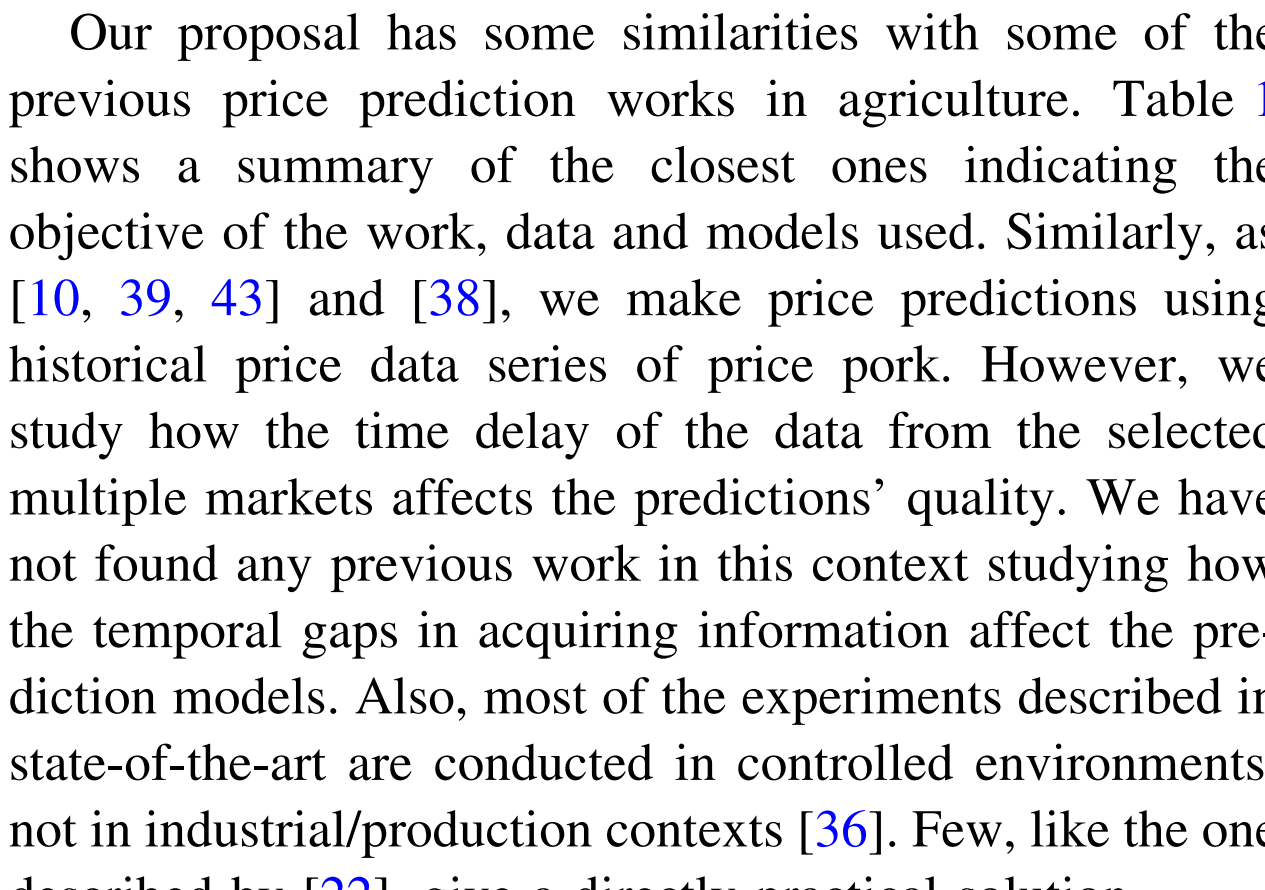
Our proposal has some similarities with some of the previous price prediction works in agriculture. Table 1 shows a summary of the closest ones indicating the objective of the work, data and models used. Similarly, as [10, 39, 43] and [38], we make price predictions using historical price data series of price pork. However, we study how the time delay of the data from the selected multiple markets affects the predictions' quality. We have not found any previous work in this context studying how the temporal gaps in acquiring information affect the prediction models. Also, most of the experiments described in state-of-the-art are conducted in controlled environments, not in industrial/production contexts [36]. Few, like the one described by [22], give a directly practical solution.

Concerning the prediction models, works such as [17, 23, 25] and [34] also provide model comparisons. Our work covers various statistical and machine-learning models suitable for sequence prediction tasks. Specifically, we have used SVM, Random forest, Ridge Regression and extremely randomized trees, which are used in many of the previous works; ARIMA and SARIMAX models that are used in [23] and [38]; LSTM and RNN models similar to the used in [7, 29]; and the more recent XGBoost, LGBM and CatBoost models used in [29] and [28]. With this collection of models, we have a broad basis for comparing results from data with the different temporal gaps in their acquisition.

## 3 Materials and methods

This section shows the data collection for training the prediction models, analyses their relations, and describes the subset selected for the experiments.

### 3.1 Data sources

In Europe, national agencies collect and report agricultural and livestock market prices to Eurostat. Related to pork meat, Eurostat provides monthly reports with the weekly prices of 22 European countries following the S.E.U.R.O.P. pork classification system.[2] This system classifies pork quality based on the percentage of lean meat in the following categories: S -"Superior" (60% or more), E - "Excellent" (55–59.9%), U - "Very good" (50–54.9%), R - "Good" (45–49.9%), O - "Fair" (40–44.9%), P - "Poor" (40% or less). Eurostat uses the "Superior", "Excellent", and "Good" values from this system.

[2] https://ec.europa.eu/info/food-farming-fisheries/farming/facts-and-figures/markets/overviews/market-observatories/meat/pigmeat-statistics_en Last visit May, 2023.

**Table 1** Similar publications

| | Year | Proposal | Data | Models |
|---|---|---|---|---|
| Wang et al. [39] | 2022 | Propose a model for hog price forecasting by using the WOA algorithm to optimize the parameters of a LightGBM model | Daily data from the Dalian Commodity Exchange, Wind database, and Huarong Rongda data analyst website. From 8 January 2021, to 22 July 2021 | LightGBM, XGBoost, GBDT and WOA-LightGBM-CEEMDAN |
| Klompenburg [36] | 2022 | Conduct a literature review on data-driven decision-making in pig farms focusing on real business cases using data from actual production chains | Science Direct, Scopus, Web of Science, Springer Link, Wiley. From 2006 to 2021 | Review |
| Chen et al. [8] | 2022 | Develop a combination of Neural Network models for pork price prediction | Data monthly pork price data provided by the China National Database, From 2003 to 2018 | RNN, LSTM, GRU, Bi-RNN, Bi-LSTM, Bi-GRU, SVM, GBDT |
| Punia and Shankar [27] | 2022 | Provide a decision-making system that uses deep learning and ensemble models for real-time demand forecasting in retail stores | Weekly sales data for 55 food items from a large retailer. From January 2009 to January 2012 | Ols, ARIMA, ARIMAX, RF, NN, LSTM, ARIMA + NN, ARIMA + RF, LSTM + RF, PCA |
| Ye et al. [42] | 2021 | Proposes a model to predict weekly hog prices using historical prices and discussions from the online professional community | Historical prices of the hog, maize, and bean. From 2013 to 2020 | LSTM, MLP, STL-ATTLSTM, BERTLSTM, GCNLSTM |
| Chuluunsaikhan et al. [10] | 2020 | Present a methodology and models that predict pork's daily retail price in the South Korean domestic market based on news articles | News articles from PigTimes, KAMIS and EKAPEPIA. Data from 2010 to 2019 | ARIMA, Ridge, Rf, GBM, MLP, CNN, LSTM |
| Zhang et al. [43] | 2020 | Use a hybrid model that suggests price intervals instead of making specific forecasts | The weekly pork price of the whole sales market in China. Wind Database. Data from January 2009 to December 2018 | SVR, ELM, Coin-SVR VECM-CoinSVR VECM-SVR Holt-CoinSVR Holt-SVR |
| Ribeiro [28] | 2020 | Propose predicting soybean and wheat prices using historical and wheat prices for short-term forecasting | Monthly prices paid to the Parana, Brazil producers of soybean and wheat. Data from 2001-2018 for soybean and 2004 to 2018 for wheat | GBM, XGBoost, RF, SVR, KNN |
| Ma et al. [22] | 2019 | Design a network service platform for real-time monitoring of each pig's feeding status, identification, and health and determines the exact quantity of feed allocated to each pig based on its feeding status and weight | Information is taken from a pig through the feeding management equipment's hardware containing a health and environmental monitoring unit | Do not apply |
| Ma et al. [23] | 2019 | Compare the performance of different price prediction models for pork using livestock prices, feedstuff, and market indexes | Data from the National Bureau of Statistics, National Data Centre, Bric Agricultural Data Terminal. Data from 2001-2016 | BN, SVR, FCN, ARIMA |
| Wang [38] | 2019 | This study details the development of an integrated cloud service agricultural platform for market tracking and a short-term forecasting model for the pig price index | Pig prices, location, number of pigs, total weight, price per unit, and trading volumes from an agricultural internet platform for Henan and Fujian provinces in China. 2016 Data | ARIMA |
| Shahinfar et al. [29] | 2019 | Propose models to infer animal features that affect pricing based on phenotypic and environmental predictor features | Live weight, carcass and environmental records from the Sheep CRC Information Nucleus Flock. Data from 2008 to 2018 | DNN, GBM, DT, RF |
| Tao et al. [34] | 2017 | Use an empirical ensemble model decomposition to extract high, low and trend frequency behavioural patterns in the evolution of hog prices | Monthly hog prices from the Ministry of Agriculture of the People's Republic of China. From January 2000 to May 2015 | ARIMA, FNN, SVR, ELM, EEMD-ARIMA, EEMD-FNN, EEMD-SVR, EEMD-ELM |

In Spain, the Ministry of Agriculture provides the price data to Eurostat, including the weekly mean price of pork meat obtained from slaughterhouse reports. With a two-week delay, the Ministry also publishes weekly reports that include prices from regional pork meat markets in Barcelona, Huesca, Zaragoza, Lleida, Murcia, Pontevedra, Salamanca, and Segovia.[3] These reports use a meat classification system compatible with the one used in Eurostat. The prices of these local markets can also be obtained the day they are formed, without the two weeks delay, by subscription (Monday: Salamanca, Zaragoza; Tuesday: Pontevedra; Wednesday: Huesca; Thursday: Murcia, Segovia, Lleida).

## 3.2 Data selection

To select the data set for predicting the Lleida price market, we have analysed the correlations among all the previously described data using Pearson correlation analysis. We have considered using the data of markets highly correlated with the Lleida market as training data for the used models since a high correlation indicates a similar expected behaviour over time. . In the data, we visually identified the presence of an outlier. A data point with a value one unit less than its neighbours in the same market. To address this issue, we replaced the outlier with the mean value between the previous and next values of that point before conducting the correlation analysis.

The correlation pairs greater than 0.90 are shown in Fig. 1. The figure shows a high correlation between all the Spanish regional markets and Portugal. Additionally, central Europe behaves as a loose group, being relevant correlations among groups of neighbouring countries. These correlations are not only caused by a similar country's socioeconomic context but also by international trading and agreements between countries [1].

To predict the prices of the Lleida market, we selected those markets with a correlation with Lleida greater than 0.98. This value is high enough to select only highly correlated markets that could anticipate the behaviour of the Lleida market. It includes all the local Spanish markets.

The selected data are visually shown in Fig. 2. They contain the prices of 322 weeks, from January 2016 to February 2022, for each selected market. The Augmented Dickey–Fuller Test (ADF test) shows that these time series are stationary. This indicates that their mean and variance are stable over time, thus allowing for more straightforward predictions [14, 18].

We have organized the data according to the two experiments performed. The first one uses the weekly public information published by the Spanish agricultural ministry, so we consider the two-week delay in the data publication for training. The second one considers that the information has been obtained from the market's subscription, so the weekly prices are available without delay. In this case, the publication day of the week of each market is relevant as we can use training data of the same week to make the following prediction. Since Lleida prices are published on Thursday, the training input uses the prices of all the markets published between the previous Thursday and Wednesday.

## 3.3 Model descriptions

This section describes the models compared in the experiments. We have selected ARIMA/SARIMAX, Random Forest, and Support Vector Regression as a baseline because of their frequent use in previous works. Autoregressive Integrated Moving Average (ARIMA) [26] is a statistical model used over a single data time series that combines AutoRegression and Moving Average models to adjust to cyclical data. SARIMAX is very similar to the ARIMA model. Still, it adds a set of autoregressive and moving average components that allow for differencing data by seasonal frequency and non-seasonal differencing [26]. Random Forest for Regression [6] is a supervised machine learning technique that builds multiple concurrent decision trees and provides an average of the result as output. Support Vector Regression [13] defines hyperplanes minimizing the distance between the data within a $\pm\mathcal{E}$ boundary and the hyperplane.

We also compare models designed as an ensemble of multiple models to improve performance. Specifically, we have selected Extremely Randomized Trees, LGBM, XGBoost, Ridge and Catboost due to the performance shown in other contexts. Extremely Random Trees [16] are similar to random forests but differ in input data and how tree nodes are split. LGBM [20] is another tree-based learning algorithm that, instead of growing the trees horizontally like other algorithms, grows them vertically. The vertical growth avoids verifying all the leaves, increasing their speed. XGBoost [9] uses a collection of weak decision trees that are improved in each iteration with the purpose of minimizing the objective function. Ridge Regression [37] is a technique for analysing multiple regression data that suffer from multicollinearity. It reduces the standard errors by adding a degree of bias to the regression estimates. Finally, CatBoost [3] builds upon the theory of decision trees and gradient boosting. Its main idea is to combine many weak models sequentially and thus, through greedy search, create a robust predictive model.

The last family of models we have considered are those using neural networks [5]. We have selected RNN [32] and LSTM [32] models for their suitability for data sequence analysis. Recurrent Neural Networks (RNNs) are created from simple neurons that retro-feed the previous and new outputs. It makes the neuron capable of reusing information about the earlier inputs but can not encode long-term details. Long Short-Term Memory Networks (LSTMs) are created to solve the limitations of RNNs. They provide a hidden state that can remember and forget things selectively, which overcomes the weight decay problem of RNNs.

[3] https://www.mapa.gob.es/es/estadistica/temas/publicaciones/informe-semanal-coyuntura/default.aspx May, 2023.

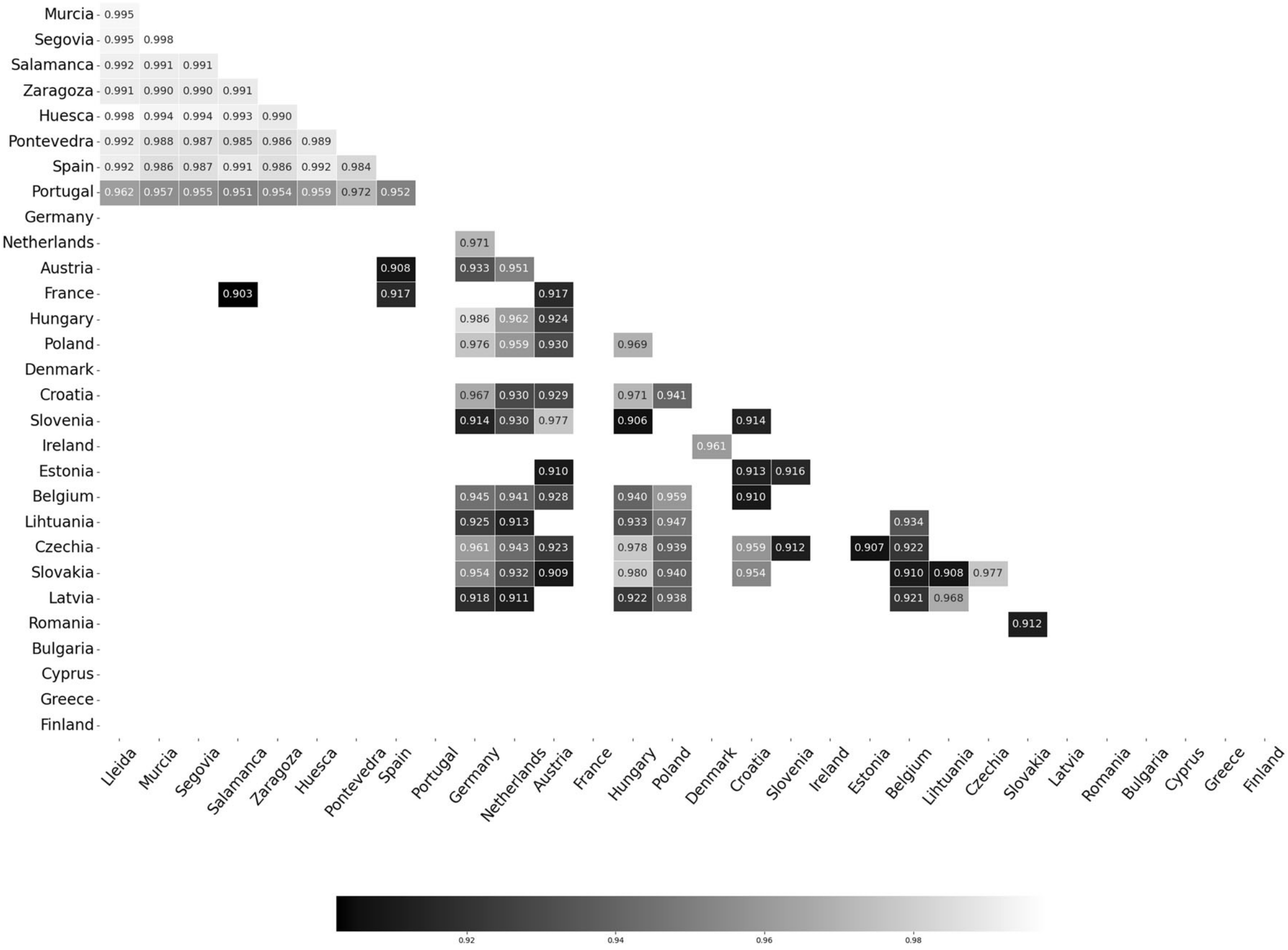

**Fig. 1** Correlation between all pork meat European markets

**Fig. 2** Pork prices in Spanish markets

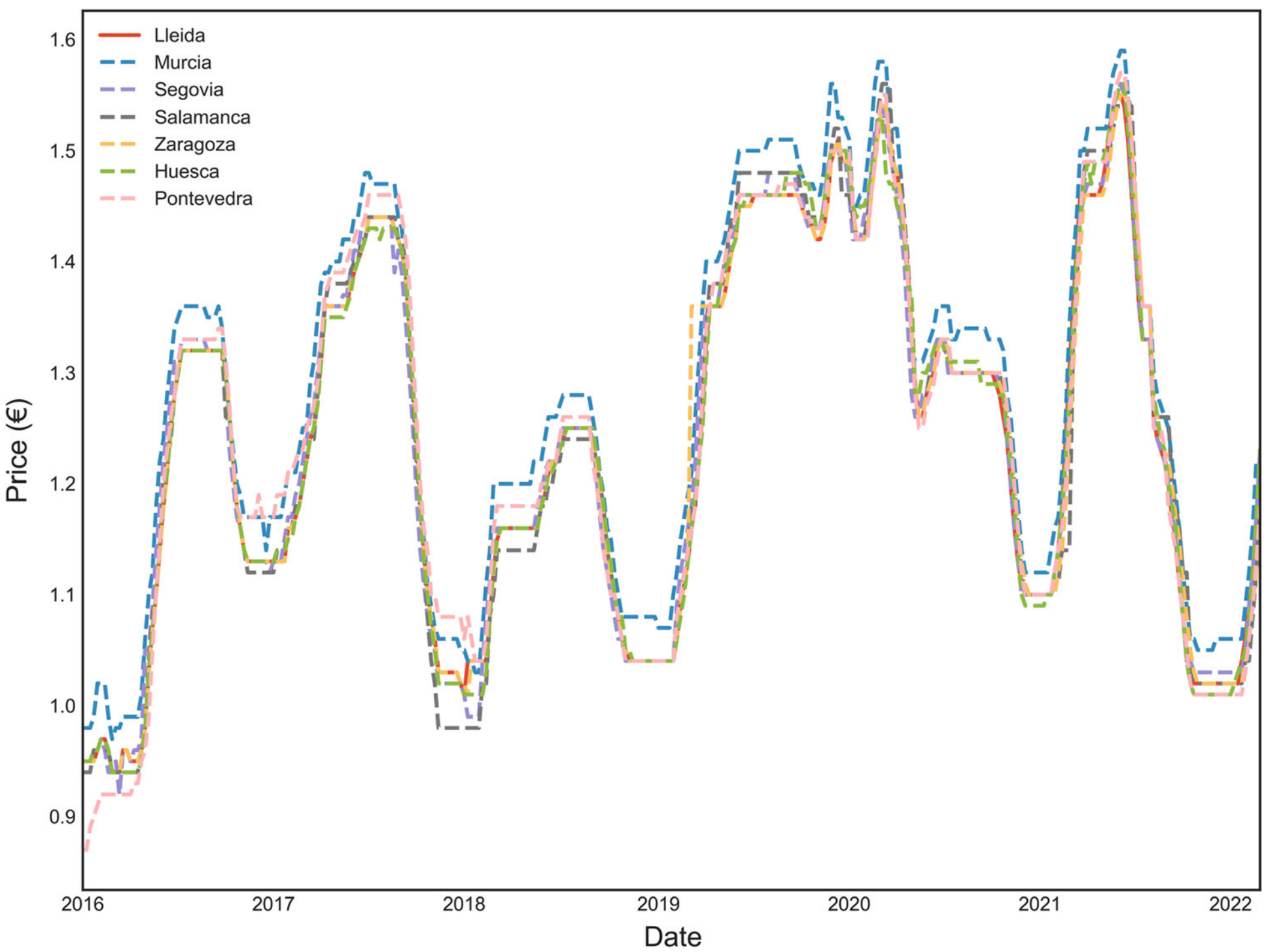

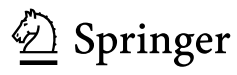

### 3.4 Model training

The data collection has been divided into training and testing in an 80:20 ratio. Since the data are a time series, we use older data points for training and the newest points for testing. All the models except Arima/Sarimax use the data of all the markets as training input to perform the prediction. Arima/Sarimax only use a single data series in their training processes. Therefore, they are trained with the data series of the Lleida market.

The models receive the data organized as n-week time windows to predict the next value. The use of time windows [15] involves partitioning a time series into smaller, overlapping subsets of data based on predetermined window size and step size, which determines the number of data points included in each window and the shift in time for each subsequent window. Our experiments study the impact of varying time window sizes on the computational cost and prediction accuracy of time series forecasting. Specifically, time windows of different lengths ranging from 2 to 12 weeks were used to identify the optimal window size that balances the trade-off between prediction accuracy and computational efficiency.

The metrics used to predict performance are the root mean square error (RMSE) and R-square (R2) calculated according to Eqs. 1 and 2. In the equations, n refers to the total number of training values, $\bar{y_i}$ is the true value, and $y_i$ is the predicted value.

$$R^2 = 1 - \frac{\sum_{i=1}^{n} (y_i - \bar{y_i})^2}{\sum_{i=1}^{n} (y_i - \bar{y_i})^{2'}} \tag{1}$$

$$\mathrm{RMSE} = \sqrt{\frac{1}{n} \sum_{i=1}^{n} (y_i - \bar{y_i})^2} \tag{2}$$

$R^2$ shows how much of the variance in the dependent variable can be explained by the independent variables. Generally, the higher the R2, the better the model fits the data. RMSE is the standard deviation of the prediction error. This provides a measure of the difference between the price predictions and the real prices of the meat that can be used to analyse the economic impact of the prediction errors. For this research, each model's hyper-parameter has been selected with OPTUNA [2], a framework that automates the hyper-parameter optimization in models of multiple machine learning libraries. Optuna uses Bayesian optimization to efficiently search the hyperparameter space and find the optimal values depending on the hyperparameters included and the ranges of values to be taken. Bayesian optimization methods seek global optimization, which is done by creating an iterative probabilistic model of the function that assigns the different values of the hyperparameters to the objective function. This model captures the function's behaviour to create a posterior distribution over the objective function and then creates an acquisition function to determine the next best probability point of improvement of the previous result. We used 1000 Trials per algorithm to obtain the best hyperparameters, and the best model was saved.

## 4 Experimental results and discussion

Table 2 shows the results of the best configuration of each model using public information (with two weeks delays) and subscription information (obtained without delays). The results are ordered by the best R2 in the results, and the RMSE metric is also included. The values of the hyper-parameters used in each model to obtain these results are shown in Table 3. This table presents the models with the best hyper-parameters and the time windows expressed in weeks of data used to train each model.

The results in Table 2 show that, independently of the organization of the input data, the Ridge model has the best performance, followed by the classic Arima/Sarimax models. For almost all the models, the results are better when there is no delay between the data publication and its availability for the model's training.

**Table 2** Comparison of model performances using public and subscription data

| Model name | Public data | | Subscription data | |
|---|---|---|---|---|
| | RMSE | R2 | RMSE | R2 |
| Ridge | 0.02041 | 0.98743 | 0.01477 | 0.99342 |
| Arima | 0.01800 | 0.98990 | 0.01800 | 0.98990 |
| Sarimax | 0.02000 | 0.98820 | 0.02000 | 0.98820 |
| Support Vector Regressor | 0.02836 | 0.97575 | 0.02134 | 0.98626 |
| XGBoost Regressor | 0.03223 | 0.96894 | 0.02254 | 0.98467 |
| LGBM Regressor | 0.03100 | 0.97101 | 0.02314 | 0.98385 |
| Random Forest | 0.03413 | 0.96486 | 0.02329 | 0.98364 |
| Extremely Random trees | 0.03451 | 0.96408 | 0.02764 | 0.97695 |
| RNN | 0.03037 | 0.97218 | 0.02972 | 0.97358 |
| LSTM | 0.03652 | 0.95978 | 0.03124 | 0.97081 |
| CatBoost Regressor | 0.03530 | 0.96241 | 0.03148 | 0.97011 |

**Table 3** Hyper-parameters used in the experiment models

| Model | Public information | | Subscription information | |
|---|---|---|---|---|
| | Week | Hyper-parameters | Week | Hyper-parameters |
| ARIMA | 4 | $p = 4$, $d = 0$, $q = 0$ | 4 | $p = 4$, $d = 0$, $q = 0$ |
| SARIMAX | 12 | $p = 1$, $d = 1$, $q = 2$, $P = 0$, $D = 1$, $Q = 1$, $M = 12$ | 12 | $p = 1$, $d = 1$, $q = 2$, $P = 0$, $D = 1$, $Q = 1$, $M = 12$ |
| Rnd. Forest | 6 | Bootstrap = True, max_features = auto, min_samples_leaf = 0.002446626, min_samples_split = 5, n_estimators = 99 | 2 | Bootstrap = True, max_features = auto, min_samples_leaf = 0.002446626, min_samples_split = 5, n_estimators = 99 |
| SVR | 4 | Degree = 6, epsilon = 0.002203, Gamma = Scale, Kernel = linear, C = 0.151861 | 2 | Degree = 6, epsilon = 0.00220298, Gamma = Scale, Kernel = linear, C = 0.151861 |
| Extr. Rnd. Trees | 5 | Bootstrap = False, criterion = squared_error, max_depth = 198, max_features = auto, min_samples_leaf = 0.012833897, min_samples_split = 2, n_estimators = 115 | 5 | Bootstrap = False, criterion = squared_error, max_depth = 198, max_features = auto, min_samples_leaf = 0.012833897, min_samples_split = 2, n_estimators = 115 |
| LGBM | 4 | learning_rate = 0.080865967, max_depth = 3, min_child_samples = 5, n_estimators = 185 | 3 | learning_rate = 0.080865967, max_depth = 3, min_child_samples = 5, n_estimators = 185 |
| XGBoost | 7 | learning_rate = 0.111361130, max_depth = 5, n_estimators = 114 | 3 | learning_rate = 0.086825338, max_depth = 5, n_estimators = 95 |
| Ridge | 2 | Alpha = 0.010034555, Solver = Cholesky | 2 | Alpha = 0.010034555, Solver = Cholesky |
| CatBoost | 3 | learning_rate = 0.218951837, max_depth = 5, n_estimators = 148 | 4 | learning_rate = 0.14997851, max_depth = 5, n_estimators = 148 |
| RNN | 7 | Neurons = 1024/256/1, Dropout = 0.02, Activation = Relu, Optimizer = Adam, Epochs = 500, Batch_Size = 10, Loss = Mean Squared error | 6 | Neurons = 1024/256/1, Dropout = 0.02, Activation = Relu, Optimizer = Adam, Epochs = 500, Batch_Size = 10, Loss = Mean Squared error |
| LSTM | 6 | Neurons = 200/100/50/1, Dropout = 0.02, Activation = Relu, Optimizer = Adam, Epochs = 500, Batch_Size = 10, Loss = Mean Squared error | 6 | Neurons = 200/100/50/1, Dropout = 0.02, Activation = Relu, Optimizer = Adam, Epochs = 500, Batch_Size = 10, Loss = Mean Squared error |

In the case of the Arima/Sarimax models, there is no difference in using public or subscription information because the model is trained a single time with only Lleida prices, and the predictions a week or two weeks forward use this same model. The fact that Arima/Sarimax models work better than most other models means that these other models cannot take advantage of the additional information provided by the rest of the markets they use as input. Only the Ridge model provides better results, and only when having subscription information. When using public data, Arima is the best model.

In terms of R2, the differences between the best model (Ridge with subscription data) and the worst (CatBoost with public data) may seem small (0.99342 vs 0.96241), but the RMSE shows an error of more than double (1.4 vs 3.5 euro cents). This has a significant economic impact, as a difference of a single cent per kilogram of meat in a farm with hundreds of animals becomes a substantial loss.

It is also important to note the number of weeks of data the models use to make the predictions. In Table 3, the column "Week" indicates the optimal time window ranging from 2 to 12 weeks. It is worth noting that the Ridge model, which provides the best results, only utilizes information from the two previous weeks to generate predictions. This suggests that there are no long-term trends that significantly affect the price changes and that only the most recent prices are sufficient for making accurate predictions.

From the hyper-parameters shown in Table 3 is important to note that the hyper-parameter configuration selected by Optuna is equal to or very similar for each model. This indicates that the models are quite independent of the training data, which is a good sign of their applicability to other similar problems or this same problem if additional training data features were available.

## 5 System prototype

To provide agricultural product producers and consumers with a reference model for conducting their business, several national governments around the world have created third-party agents that act as brokers between sellers (usually farmers) and buyers (such as slaughterhouses, food processing companies, and supermarket chains) to propose reference prices. In Spain, these third parties are known as “Lonjas” and are subject to national regulations. A “Lonja” brings together representatives from various sectors involved in the buying and selling a specific agricultural product (such as cereals, pork meat, or red berries) on a regular basis, typically weekly. They agree on a reference price that can be used as a starting point for negotiations in B2B markets during the following period. Although the price is not mandatory, it serves as a widely recognized reference point.

The price-establishing process begins with each agent submitting their initial proposal separately. Then, a technician from the “Lonja” combines the proposals to obtain a view from both sides. The agents are brought together in the same space, where the technician explains the different viewpoints and proposes a price. If all parties agree with the proposal, it becomes the reference price. If not, a discussion starts to justify the different proposals presented. The process ends when an agreement is obtained. In some cases, if an agreement cannot be reached, the technician must decide on the price.

Similar to international stock markets, the importance of each “Lonja” varies, and different farm products have their own reference “Lonjas”. Usually, each region/country has a reference agent and others that complement them along the week or according to special local situations. For example, in Spain, the main reference “lonja” for pork is Mercolleida[4] in a medium-sized city near Barcelona. Meanwhile, the most relevant one for bovine is “Lonja” Binéfar[5] in a small village in the northeast of Spain.

Since 2018, Lonja Binéfar has been working to establish itself as an alternative option for pork reference prices, in competence with Mercolleida. To achieve this, we have been working with this entity to give them a technological tool that could complement their decision process. The idea is to have a third option that can be compared with the proposal from both sides. This third option is based on the model presented in the previous section and is only used if the decision-makers involved in the process disagree.

The system core is a web application that, every week, download data from Spanish and European markets and publish them in a web portal,[6] showing the data collected from the different markets (see Fig. 3).

The collected data are provided to the prediction model that shows the technician the following week’s price proposals and if it corresponds with an increase or decrease in the price. The prediction model used is the best-performing model from the previous section. However, it is trained to predict prices from “Lonja” Binéfar instead of Mercolleida, which has been used in the experiments as the reference pork market in Spain. Figure 4 compares the proposals made by the prediction system and the price agreed upon in the “Lonja” process from May 2022 till the middle of October 2022.

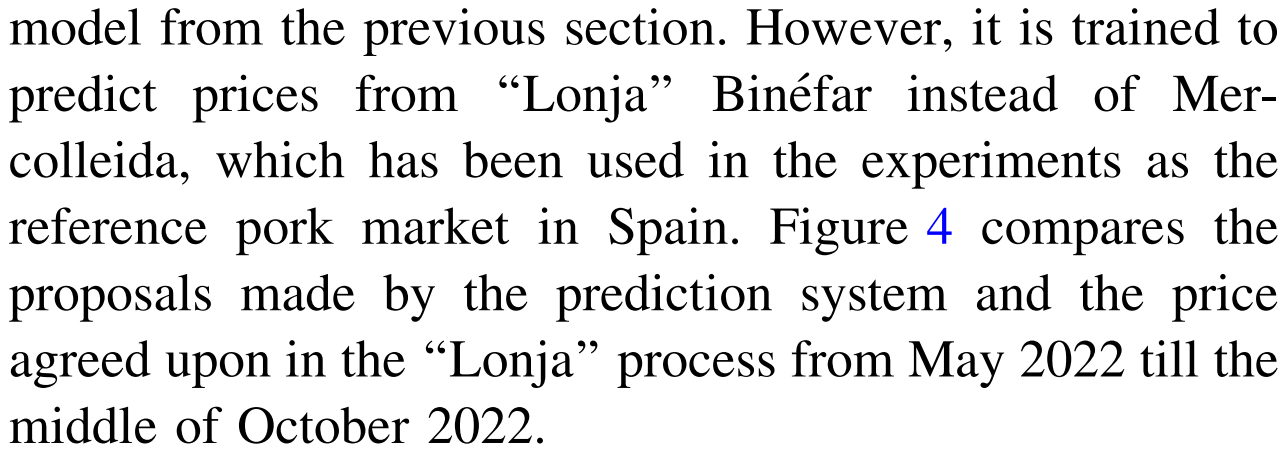

The maximum difference between values in a week has been 0,02 €, which implies an error of 1,2 %. It has been considered a good result by “Lonja” Binéfar (considering the pork market’s current volatility).

## 6 Conclusions and future work

In this paper, we analyse the impact of time lags in acquiring historical data on the performance of a prediction system. We have used the available market information about pork prices in European countries and compared different machine learning models for predicting the price of pork. In the experiments, we focus on the Spanish local market of Lleida and study the effect of the time gap of the available data in price prediction. This is done by comparing the results of using public data with a delay of two weeks or subscription data that can be obtained the moment it is decided.

A high correlation has been found between the markets of Spain and Portugal, but the strongest correlations have been among Spanish local markets. Other correlations have been found among groups of central Europe countries, which indicates that the European pork market does not work as a unity but as a group of independent region markets.

The experiment results have shown that even though all the analysed models provide reasonable predictions, the RMSE varies significantly between solutions. The model providing the best solution is Ridge using subscription data, with a RMSE of 1.4 cents. When using public data, this error increases to 2 cents. The worse solution has an error of up to 3.5 cents. Economically, this error difference makes using the model with the best performance relevant.

Another important aspect identified is that using longer data sequences to train the models does not mean better results, and many models generate noise that worsens the results. Ridge best solution uses data of only two weeks to make the prediction, and generally, the best data interval is between 2 and 4 weeks.

Finally, the development of the decision system has shown the applicability of the proposed solution to a real case scenario, providing approximate price values to the market behaviour in the Binefar market.

Future work will extend this work by considering additional elements in the predictions. Specifically, we want to include the opinion of stakeholders, information on the market state from technical reports or newspapers and market prices of products related to pork production, such as corn, oats or wheat.

[4] https://www.mercolleida.com Last visit May 2023.

[5] http://www.lonjabinefar.es Last visit May 2023.

[6] https://www.preciolonja.es Last visit May, 2023.

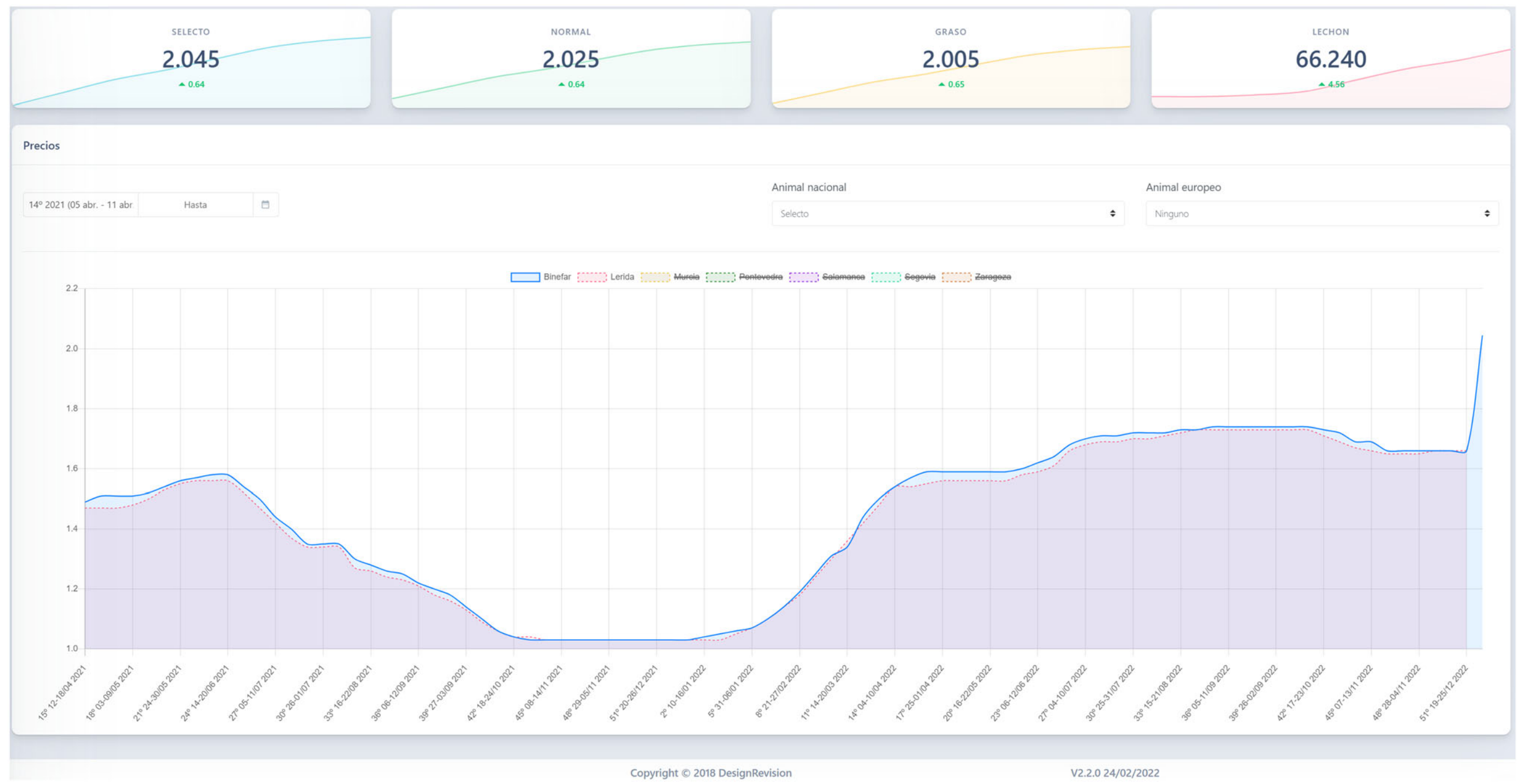

**Fig. 3** Prototype main page

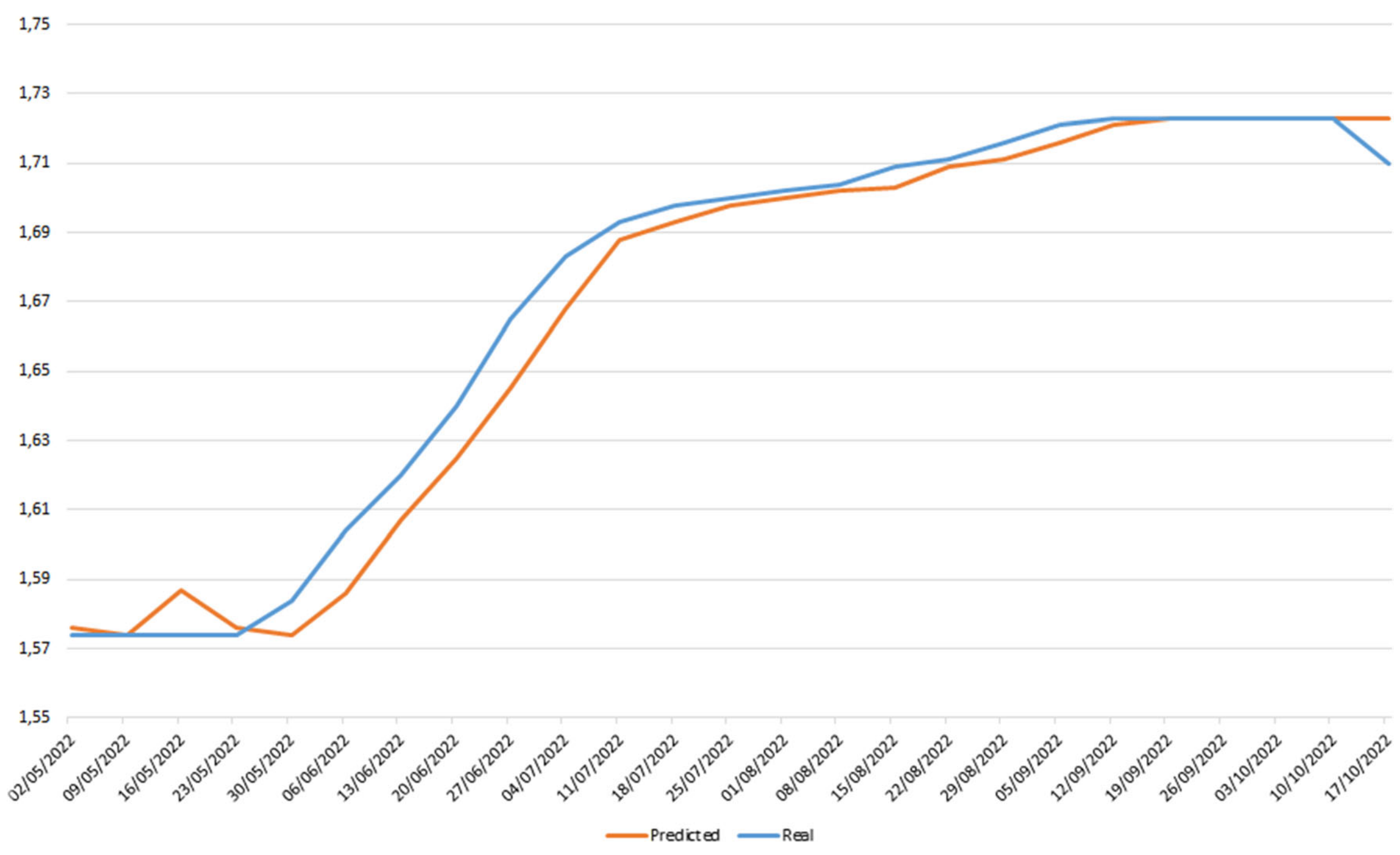

**Fig. 4** Comparing Lonja's proposal and the predicted proposal

**Funding** Open Access funding provided thanks to the CRUE-CSIC agreement with Springer Nature. This work has been partially supported by the Aragon Regional Government and the European Union - FEADER (projects GCP2019006700 and T59_23R), the Spanish Government (project PID2020-113353RB-I00). The work of Mario

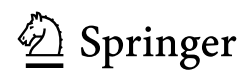

Suaza has been partially supported by the Colombian Ministry of Science, Technology and Innovation (MINCIENCIAS 885/2020).

**Data availability** The data that support the findings of this study are publicly available online at Pigmeat statistics Europe and the weekly economic outlook in Spain.

## Declarations

**Conflict of interest** The authors declare that they have no conflict of interest.

**Ethical approval** This article does not contain any studies with human participants or animals performed by any of the authors